\documentclass[acmsmall, nonacm]{acmart}

\usepackage[most]{tcolorbox}
\usepackage{listings}

\AtBeginDocument{%
  }

\begin{document}

\title{Token-Level Adaptation of LoRA Adapters for Downstream Task
  Generalization}

\author{Joshua Belofsky}
\affiliation{%
  \institution{University of Chicago}
  \city{Chicago}
  \state{Illinois}
  \country{USA}}
\email{jb1@uchicago.edu}

\renewcommand{\shortauthors}{}

\begin{abstract}
  This paper introduces a method for adapting LoRA adapters in smaller-sized language models to arbitrary downstream tasks. Unlike standard mixture-of-expert architectures, our method employs a gradient-free routing function to choose a weighted combination of experts without increasing the compute requirements for training or inference. The results show that token-level adaptation of LoRA adapters outperforms the base Llama-2-7b model across mathematical (GSM8K), scientific (ARC-Challenge), reading comprehension (SQuAD), and coding (CodeAlpaca-20k) tasks. Further evaluations also show that the average performance of token-level adaptation outperforms individual models fine-tuned for each of the tasks with the best performance observed in adaptation of every-other token during inference. The code for this study is made available through a public repository.\footnote{\url{https://github.com/jb-01/LoRA-TLE}.}
\end{abstract}

\ccsdesc[500]{Computing methodologies~Machine learning approaches}
\ccsdesc[500]{Computing methodologies~Neural networks}
\ccsdesc[500]{Computing methodologies~Natural language processing}
\ccsdesc[300]{Mathematics of computing~Nonconvex optimization}

\keywords{Large Language Models, Low-rank Adaptation, Mixture of experts, Downstream task generalization}


\maketitle

\section{Introduction}
Large language models (LLMs) excel at a broad range of tasks, thanks to extensively pre-training on vast datasets \cite{brown2020language,radford2018improving}. The effectiveness of these models is attributable to both the scale of the model architecture and the size of the training data \cite{kaplan2020scaling,raffel2020exploring,hestness2017deep}. These advancements, along with algorithmic improvements such as attention mechanisms, mark a significant departure from earlier, smaller neural networks that often suffered from catastrophic forgetting when trained on disparate tasks \cite{vaswani2017attention}.

This paper implements a method enabling smaller sized language models (7B parameters) to generalize across a spectrum of downstream tasks. We use low-rank adaptation (LoRA) to freeze the pretrained model weights and inject trainable rank decomposition matrices which can provide a parameter-efficient alternative to full fine-tuning \cite{hu2021lora}. Parameter-efficient fine-tuning, including LoRA, are capable of performance at or above the levels of larger-sized language models in specific domains \cite{peft1,peft2,llama-adapters,dettmers2023qlora}. However, the performance of these models is limited to the domain for which they are fine-tuned. This paper introduces a method for context-aware adaptation of LoRA adapters to different domains, enabling enhanced performance across mathematical, scientific, reading comprehension, and coding tasks. We demonstrate that our token-level adaptation approach not only outperforms the base Llama-2-7b model but also achieves better average results than models fine-tuned for individual tasks, particularly when adapted at a frequency of every other token.

\section{Background}
Achieving human-level proficiency in reasoning, mathematics, reading, and language has been greatly advanced by LLMs. However, attaining state-of-the-art results across multiple domains typically requires significant computational resources and extensive pre-training \cite{openai2023gpt4}. To address this, a common approach is to concentrate efforts on a specific area, fine-tuning smaller models with domain-specific data. Parameter-Efficient Fine-Tuning (PEFT) allows smaller models to achieve performance comparable to larger counterparts by focusing on targeted domains. To increase the generality of these fine-tuned models, a Mixture-of-Experts (MoE) architecture can be used, which integrates various specialized models into a single unit. Our paper seeks to combine the strengths of both PEFT and MoE frameworks by utilizing LoRA adapters to achieve efficient task generalization across four different domains.

\subsection{Low-rank Adaptation (LoRA)}
Proposed by Hu et al. in 2021, LoRA introduces a method for the efficient fine-tuning of pre-trained models using a minimal set of additional trainable parameters. By incorporating trainable low-rank decomposition matrices into pre-existing LLM layers, LoRA enables these models to adapt to new datasets while preserving their original weights. This adaptation occurs through layer-wise reparameterization, represented by the insertion of low-rank matrices in matrix multiplication operations, thus eliminating the need for recalculating dense matrices during fine-tuning.

LLM adapters are essentially compact neural modules embedded within the model, possessing a limited number of extra trainable parameters. These adapters allow for efficient task-specific fine-tuning, altering only their parameters while keeping the model's core pre-trained weights (\( \Theta \)) unchanged. This approach ensures the retention of fundamental representations learned by the LLM, with the adapters acquiring the ability to encode nuances specific to each task.

\subsection{Mixture-of-Experts}
The Mixture-of-Experts (MoE) paradigm, conceptualized in the 1990s \cite{jacobs1991adaptive,jordan1994hierarchical}, consists of several specialized sub-networks or `experts' (\(E_1, ..., E_n\)). These experts are selectively activated by a gating mechanism (\(G\)) in response to different inputs. Though large-scale MoE models have shown exceptional performance \cite{shazeer2017outrageously,fedus2022switch}, they come with significant computational overhead during both pretraining and inference.

In this paper, we implement an MoE framework that uses LoRA adapters to constitute cost-efficient domain-specific experts. Our routing function combines each individual LoRA adapter into a single weighted expert adapter using cosine similarity of the context to the centroid of each dataset. Passing the input tokens through a single expert adapter eliminates the need to compute each expert's output for every token, reducing the computational requirements for generating next-token probabilities.

\section{Token-level Adaptation of LoRA Adapters}
This section first discusses the mechanics of next-token prediction in autoregressive LLMs. Using this concept, it becomes possible to understand how the proposed method for token-level adaptation of LoRA adapters facilitates downstream task generalization.

\subsection{Next-Token Prediction}
LLMs are trained to sequentially predict next-token probabilities from all preceding tokens as input. This autoregressive decoding process is formalized as follows:
\begin{align}
  \begin{split}
    P(\mathbf{x}) &= P(x_1) \cdot P(x_2|x_1) \cdot P(x_3|x_1x_2) \cdot \ldots \cdot P(x_i|x_1x_2\ldots x_{i-1} ; \theta)
  \end{split} \\
   & = \prod_{i=1}^{n} P(x_i|x_1x_2\ldots x_{i-1} ; \theta) \nonumber
\end{align}
where $\mathbf{x} = (x_1, x_2, \ldots, x_n)$ is a sequence of tokens, and $P(x_i|x_1x_2\ldots x_{i-1})$ is the probability of the $i^{th}$ token given all preceding tokens and the model parameters, $\theta$. The model is trained to minimize the negative log-likelihood of the target sequence $\mathbf{x}$:
\begin{equation}
  \mathcal{L} = -\sum_{t=1}^{n} \log P(x_t | x_{<t}; \theta)
\end{equation}
where $x_t$ is the $t^{th}$ token in the sequence, $x_{<t}$ represents all tokens preceding $x_t$ in the sequence, and $\theta$ denotes the model parameters. This loss function quantifies how well the model's predicted probability distribution aligns with the actual token sequence $\mathbf{x}$. After pre-training, $\theta$ is a static representation of the model's learned knowledge, enabling the generation of new sequences from arbitrary prompts.

\subsection{Proposed Method}
We propose a method that dynamically combines four separate LoRA adapters in the Llama-2-7b base model \cite{llama2} based on the embeddings of the input prompt. These adapters are fine-tuned for distinct tasks: mathematics (gsm8k) \cite{cobbe2021training}, scientific reasoning (AI2\_ARC-Challenge), coding (CodeAlpaca-20k)\footnote{Dataset available at: \url{https://huggingface.co/datasets/sahil2801/CodeAlpaca-20k}}, and reading comprehension (SQuAD) \cite{2016arXiv160605250R}. The cosine similarity between the embedding of the input prompt, denoted by $\mathbf{p}$, and the centroids of the embedded datasets for each adapter ($\mathbf{a}$) is computed. These similarity scores are used to assign weights to each adapter's contribution toward predicting the next token. The most similar adapter's weight is multiplied by four\footnote{Testing indicated that a multiplier of four enhances response quality compared to unweighted similarity scores. Response quality deteriorates with multipliers of five or more.} to increase its influence. The final prediction for the next token, given the sequence of previous tokens, is derived from a weighted softmax function applied to the outputs of the adapters:

\begin{equation}
  P(x_i|x_1x_2\ldots x_{i-1}) = P(x_i|x_1x_2\ldots x_{i-1}; \theta_{\text{expert}})
\end{equation}
where

\begin{equation}
  \theta_{\text{expert}} = \sum_{j=1}^{4} w_j \cdot \theta_j
\end{equation}

and \( w_j \) is the softmax-normalized weight for the \( j^{th} \) adapter, calculated as:

\begin{equation}
  w_j = \frac{\exp(s_j \cdot T)}{\sum_{k=1}^{4} \exp(s_k \cdot T)}
\end{equation}

\begin{figure}[htbp]
  \centering
  \includegraphics[width=1\linewidth]{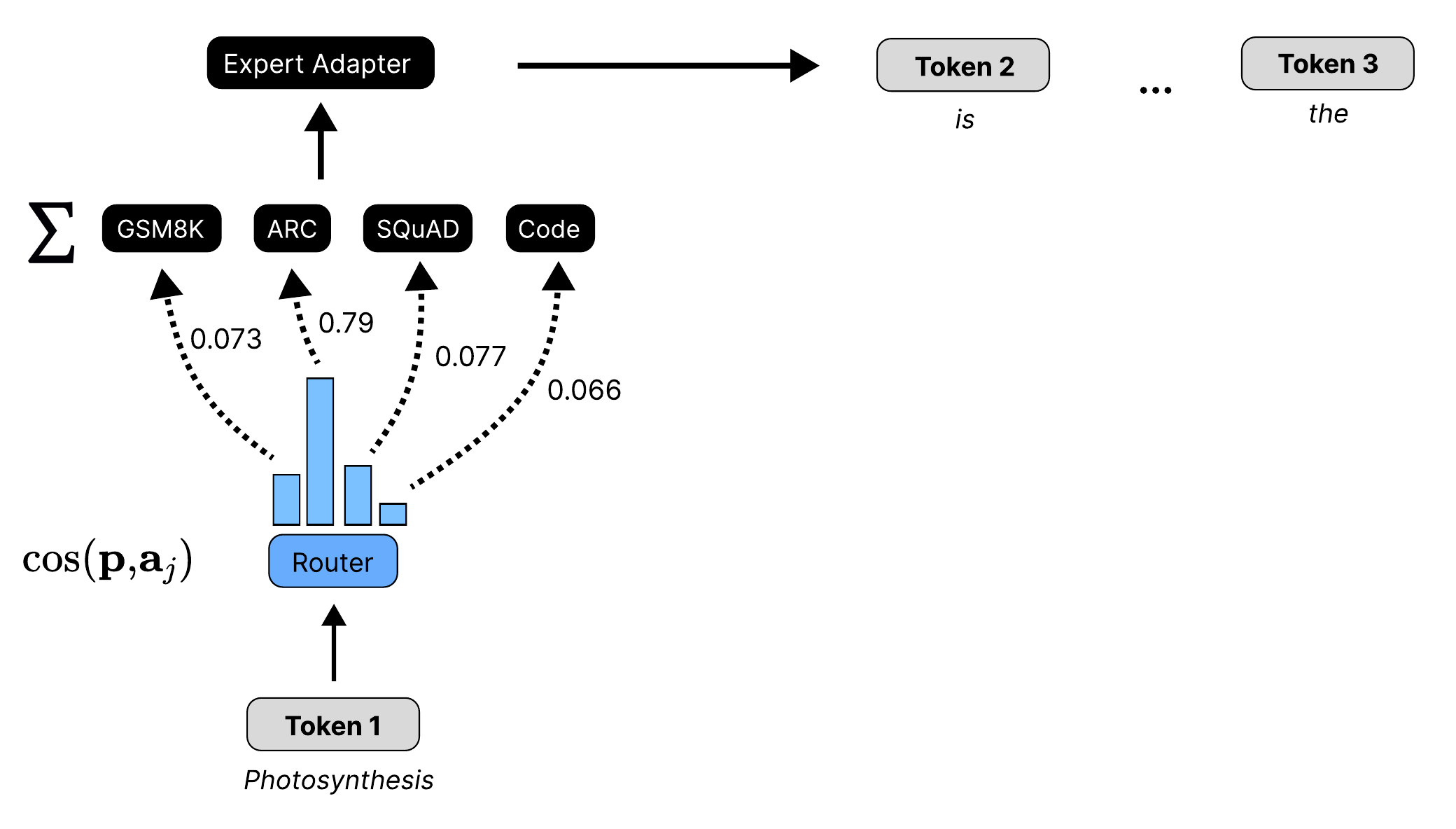}
  \caption{The architecture of the proposed method with four separate LoRA adapters for different downstream tasks.}
  \label{fig:proposed_method}
\end{figure}

Here, \( s_j = \cos(\mathbf{p}, \mathbf{a}_j) \) is the cosine similarity between the prompt embedding and the \( j^{th} \) adapter's dataset centroid embedding, \( T \) is a temperature parameter that adjusts the concentration of the softmax distribution—set to 1 for all adapters except for the most similar one, which is set to 4—and \( \theta_j \) are the parameters of the \( j^{th} \) adapter. Figure \ref{fig:proposed_method} illustrates the architecture of the proposed method.

The proposed method is inspired by the work of Fedus et al. (2022), who introduced a gradient-free routing function for MoE models. The output computation of their routing function is the linearly weighted combination of each expert’s computation on the token by the gate value. The important distinction in our method is that each token is only routed through a single expert adapter which itself is a linearly weighted combination of the previous adapters. This allows for a lightweight routing function that does not require the computation of each expert's output for every new token. Furthermore, PEFT allows for efficient fine-tuning of each individual adapter without the need to pre-train clustered adapters and perform inference over a branched network \cite{btm,c-btm}. This results in an efficient routing function that can be used to adapt to different domains without increasing pre-training requirements.

Every new token is generated using a unique weighted combination of the original four adapters on a token-by-token basis. The resulting behavior of this architecture is qualitatively different from individually activating any single set of weights fine-tuned on each of the four tasks. For example, when writing a Python function with comments, our model flexibly uses elements of both coding and literature in its output. When generating code-specific tokens, it predominantly utilizes the CodeAlpaca-20K adapter. However, when generating comments within the code, the reading comprehension adapter (SQuAD) takes precedence. This adaptive approach ensures that the resulting code maintains high quality, as it leverages the strengths of the code-specific adapter for programming syntax and the reading comprehension adapter for articulate comments. This contrasts with a model using a static set of weights, which would merely attempt to balance both tasks, potentially leading to suboptimal outcomes in both code quality and comment clarity.

\section{Experiments and Results}
This section details the experiments conducted to evaluate the efficacy of token-level adaptation of LoRA adapters in the Llama-2-7b model. We compared the performance of our method against both the base Llama-2-7b model and models fine-tuned for specific tasks.

\subsection{Experimental Setup}
We fine-tuned the Llama-2-7b model using LoRA on four datasets: GSM8K, ARC-Challenge, CodeAlpaca-20k, and SQuAD. This process yielded four specialized adapters, namely llama-2-7b-gsm8k, llama-2-7b-ai2-arc, llama-2-7b-CodeAlpaca-20k, and llama-2-7b-SQuAD. Each adapter was fine-tuned on the training dataset to optimize performance on its respective narrow task.

\subsection{Methodology}
Our methodology involved three primary modes of evaluation:
\begin{enumerate}
  \item \textbf{Base Model Comparison:} The base Llama-2-7b model was evaluated across all four datasets to establish a baseline performance metric.
  \item \textbf{Domain-Specific Benchmarking:} Each of the fine-tuned models (llama-2-7b-gsm8k, llama-2-7b-ai2-arc, llama-2-7b-CodeAlpaca-20k, and llama-2-7b-SQuAD) was evaluated on its respective domain-specific task to establish specialized performance metrics.
  \item \textbf{Token-Level Adapter Framework:} The performance of the token-level adaptation approach was assessed. In this framework, the contribution of each LoRA adapter was dynamically adjusted based on the context of the input prompt, offering a direct comparison to the base model and the domain-specific benchmarks.
\end{enumerate}

Additionally, we investigated the impact of varying the adaptation frequency of the token-level adapter. The adapter was tested at intervals of every token, every other token, every third token, and every fourth token.

\subsection{Performance Metrics and Evaluations}
To measure performance, we selected sixty questions from the test splits of each dataset. The correctness of the models' responses was evaluated, considering only the first full answer in cases of multiple outputs. Each response was manually graded as either being `correct' or `incorrect', without the option for partial credit, to maintain consistency in evaluation across different domains.

\subsection{Results}

\begin{table}[h!]
  \centering
  \begin{tabular}{l|c|c|c|c|c|c}
    \toprule
    \textbf{Model}             & \textbf{Avg.} & \textbf{ARC-Challenge} & \textbf{GSM8K} & \textbf{CodeAlpaca-20k} & \textbf{SQuAD} \\
    \midrule
    Llama-2-7b (Base)          & 16.7          & 33.3                   & 0.00           & 6.67                    & 26.7           \\
    Specialized                & 40.0          & 26.7                   & \textbf{26.7}  & 26.7                    & \textbf{80.0}  \\
    Token-level Adaptation (1) & 36.7          & 53.3                   & 6.67           & 46.7                    & 40.0           \\
    Token-level Adaptation (2) & \textbf{48.3} & \textbf{73.3}          & 6.67           & \textbf{53.3}           & 60.0           \\
    Token-level Adaptation (3) & 38.3          & 66.7                   & 20.0           & 13.3                    & 53.3           \\
    Token-level Adaptation (4) & 36.6          & 53.3                   & 6.67           & 13.3                    & 73.3           \\
    \bottomrule
  \end{tabular}
  \caption{Performance scores of each model across all four datasets.}
  \label{table:performance_comparison}
\end{table}

The results, summarized in Table \ref{table:performance_comparison}, show a comparative analysis of the average and individual dataset performances of each model. Notably, token-level adaptation demonstrated superior performance compared to the base model across all domains. It also closely matched, and in some cases exceeded, the performance of the specialized adapters.

\begin{figure}[htbp]
  \centering
  \includegraphics[width=1\linewidth]{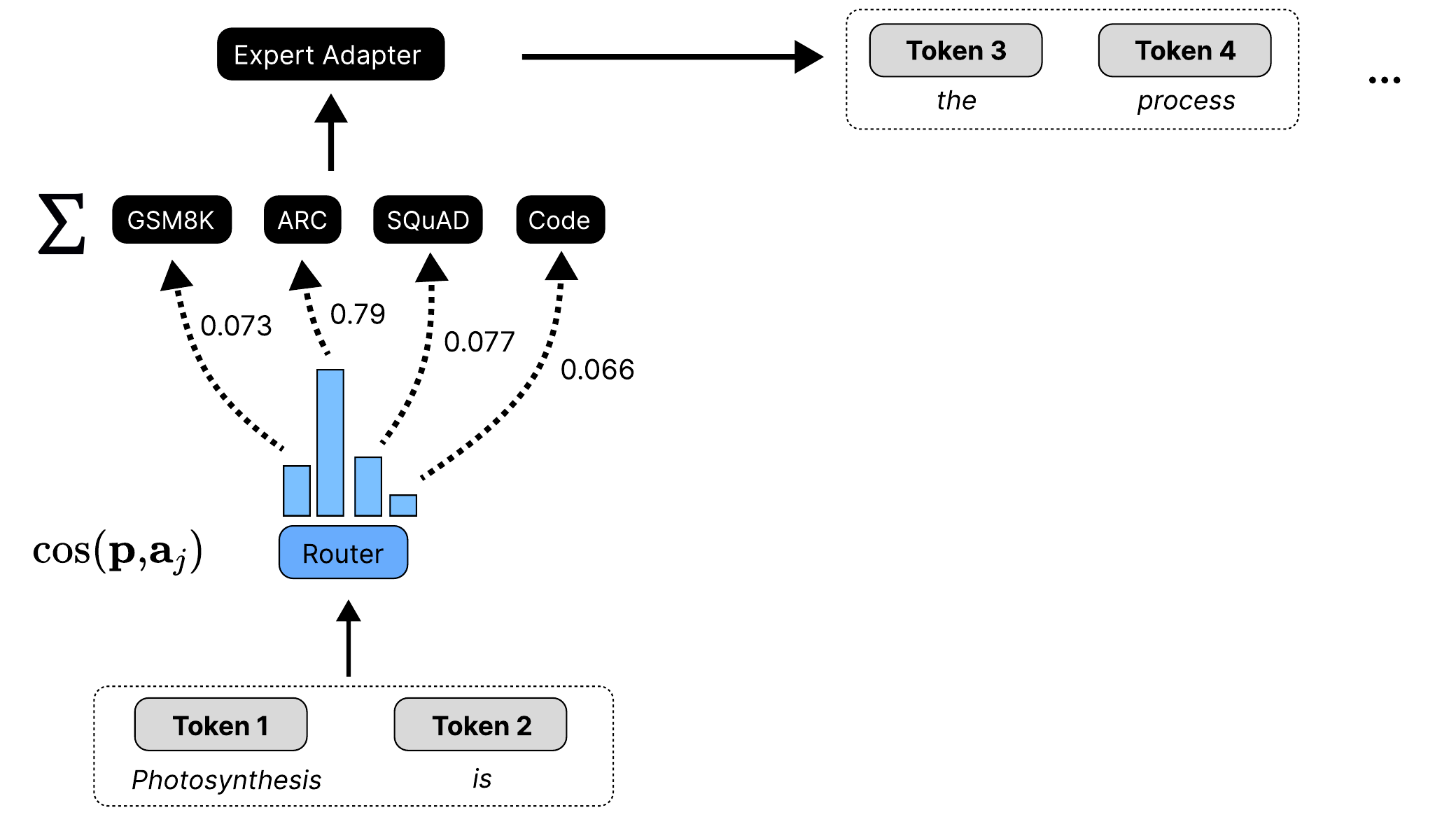}
  \caption{The highest average performing method recalculates the expert adapter for every two tokens in the context.}
  \label{fig:2-token-level-adaptation}
\end{figure}

Figure \ref{fig:2-token-level-adaptation} highlights the most effective adaptation strategy, showing that recalculating the expert adapter every two tokens yielded the highest average performance (48.3\%). This strategy also outperformed the specialized adapters in the ARC-Challenge and CodeAlpaca-20k domains.

Our findings indicate that token-level adaptation of LoRA adapters not only enhances the model's ability to generalize across various tasks but also optimizes computational efficiency by not requiring the computation of each expert’s output for every token.

\begin{tcolorbox}[
    enhanced,
    boxrule=0.5pt,
    colback=white, 
    colframe=black, 
    sharp corners,
    boxsep=5pt, 
    left=5pt, 
    right=5pt, 
    top=5pt, 
    bottom=5pt 
  ]
  \textbf{Prompt \#1:}

  Question: The planets in the solar system revolve around the Sun because the Sun is
  Choices:
  A. made of hydrogen and helium.
  B. older than other objects in the solar system.
  C. the source of most heat and light energy.
  D. the strongest source of gravity in the solar system.

  Answer: \\
  \textbf{Token-level expert (\textbf{Correct}):}

  Answer: D. the strongest source of gravity in the solar system.

  \textbf{Base Llama-2-7b:}

  Answer: B. older than other objects in the solar system.

  \tcblower
  \small{Note: This example illustrates model responses to an ARC-Challenge test question using the Llama-2-7b base model and token-level adaptation of LoRA adapters.}
\end{tcolorbox}

In the provided example, the token-level expert correctly identifies the primary reason planets revolve around the Sun (due to it being the strongest source of gravity), whereas the base Llama-2-7b model incorrectly attributes this to the Sun's age. This disparity underscores the enhanced accuracy and nuanced understanding achieved through token-level adaptation.

Building on the previous insights, we now examine another instance where the token-level expert and a fine-tuned (specialized) model respond to a Java programming question. This example further illustrates the differences in the effectiveness and accuracy of these models, particularly in technical or domain-specific contexts.

\begin{tcolorbox}[
    enhanced,
    boxrule=0.5pt,
    colback=white, 
    colframe=black, 
    sharp corners,
    boxsep=5pt, 
    left=5pt, 
    right=5pt, 
    top=5pt, 
    bottom=5pt 
  ]
  \textbf{Prompt \#2:}

  Question: Write a method `addString' in Java that takes two strings as parameters and returns a single string that concatenates the two strings.

  Answer: \\
  \textbf{Token-level expert (\textbf{Correct}):}

  \begin{verbatim}
public static String addString(String str1, String str2) {
  return str1 + str2;
}

System.out.println(addString("Hello", "World"));
// Output: HelloWorld
\end{verbatim}

  \textbf{Fine-tuned model:}

  \begin{verbatim}
public String addString(String str1, String str2) {
  return str1 + str2;
}

System.out.println(addString("Hello", "World"));

public class AddString {
  public static void main(String[] args) {
    String str1 = "Hello";
    String str2 = "World";
    System.out.println(addString(str1, str2));
  }
}
\end{verbatim}

  \tcblower
  \small{Note: This example demonstrates model responses to a CodeAlpaca-20k test set question, comparing the output of a token-level expert and a fine-tuned (specialized) model.}
\end{tcolorbox}

In this programming example, the token-level expert demonstrates its capability by correctly writing a method for string concatenation. Its response adheres to Java syntax and best practices, employing a static method to effectively concatenate two strings. This solution is both syntactically correct and logically sound, reflecting the model's ability to accurately interpret and respond to coding-related queries.

Conversely, the fine-tuned model's response, while addressing the task, is flawed with technical inaccuracies. The method it proposes is non-static, yet it is called as if it were static, which would lead to a compilation error in Java. Furthermore, the method call is placed outside the class or any method body, violating Java's structural requirements. These errors not only render the code non-functional but also indicate a lack of understanding of Java's fundamental principles.

\section{Conclusion}
Token-level adaptation of LoRA adapters outperforms the Llama-2-7b base model across mathematical, scientific, reading comprehension, and coding tasks. Further evaluations show that token-level adaptation of the expert adapter for every other token achieves better average results than models fine-tuned for each of the tasks. Our method provides an efficient way to generalize across different downstream tasks using context-aware adaptation of LoRA adapters.

These results suggest a promising avenue for enhancing the parameter efficiency and domain generalization of LLMs. Context-aware adaptation of LoRA adapters produces a better performing LLM without increasing the total parameter size or compute required for a single forward-pass of the entire network. In low-latency scenarios, such as cloud computing applications, employing smaller-sized LLMs with token-level adaptation could handle a broader array of tasks more accurately than base models, without incurring additional costs or latency.

Future research could extend this framework to even more varied tasks, potentially exploring unsupervised domains or low-resource data. Another prospective direction could involve a more complex routing function to further optimize the selection and weighting of individual adapters. The scalability of token-level adaptation could be probed with models of different sizes, investigating the upper and lower bounds of model complexity where this approach remains effective. Finally, the impact of varying the number and size of the adapters could be explored, potentially revealing the optimal limit to the number of adapters for a given model size.

\bibliographystyle{ACM-Reference-Format}
\bibliography{sample-base}

\end{document}